\documentstyle[12pt,twoside,epsf]{kaeog}
\begin{document}
\author{Marek W.~Gutowski\\
	Institute of Physics,
	Polish Academy of Sciences\\
	02--668 Warszawa, Al. Lotnik\'ow 32/46,
	Poland\\
	e-mail: gutow@ifpan.edu.pl
}

\title{AGING, DOUBLE HELIX AND SMALL WORLD PROPERTY IN~GENETIC
ALGORITHMS}

%%%%% authors  %%%%%%%%%%%%%%%%%%%%%%%%%%%%%%%%%%%%%%%%%%%%%%%%%%%%%%%%%

\maketitle

%%%%% abstract  %%%%%%%%%%%%%%%%%%%%%%%%%%%%%%%%%%%%%%%%%%%%%%%%%%%%%%%%

\begin{abstract}
Over a quarter of century after the invention of genetic algorithms and
miriads of their modifications, as well as successful implementations,
we are still lacking many essential details of thorough analysis of
it's inner working.  One of such fundamental questions is {\em how many
generations do we need to solve the optimization problem?\/} This paper
tries to answer this question, albeit in a fuzzy way, making use of the
double helix concept.  As a byproduct we gain better understanding of
the ways, in which the genetic algorithm may be fine tuned.

\end{abstract}

%%%%% text  %%%%%%%%%%%%%%%%%%%%%%%%%%%%%%%%%%%%%%%%%%%%%%%%%%%%%%%%%%%%


\section{Introduction}
The carrier of genetic information in the Nature, the DNA molecule, has
very peculiar structure.  This is a spiral-staircase shaped object,
whose steps are made from pairs selected from among only four other
kinds of molecules. Those are: cytosine (C), guanine (G), adenine (A)
and thymine (T).  There are only two correct kinds of those pairs, not
${4\choose 2}=6$, as one might naively suspect.  Namely, the cytosine
always forms a pair with guanine (\mbox{C $\leftrightarrow$ G}), and
adenine ``likes'' thymine (\mbox{G $\leftrightarrow$ T}).  So, if the
sequence in a part of DNA molecule is something like
$$
\ldots ATTGTCCA \ldots
$$
then the other half of DNA's double helix has the structure
$$
\ldots TAACAGGT \ldots
$$
Such a redundancy has not yet been exploited by genetic algorithms, at
least the present author failed to find in available literature
anything similar in conjunction with this class of optimization
algorithms (or any other class \dots).

In this paper we will show, that the idea of doubled (or duplicated)
genetic information may be very useful in fine-tuning of genetic
algorithms and in formulating the stopping criteria, which are quite
general, independent of the problem under study.

\section{The role of mutations}
Many elements of existing genetic algorithms are straightforward
computer implementations of the natural evolutionary phenomena.  We
have an evolving population, usually fixed in size, in which
individuals mate, have offspring and are mutated.  All those processes
are driven by one or more random number generators, that is by purely
stochastic forces, and, additionally, by the Darwinian rule of {\em
survival of the fittest\/.}

It is well known that the crossover processes alone cannot guarantee
finding the optimal region in a search space, at least when the
population is small.  Without mutations, the {\em premature
convergence\/} is then almost certain, thus revealing the inability of
the algorithm to find the desired solution.  The mutations cause rapid
changes of location in the search space thus making possible to reach
an explore the regions, which are at all not accessible without them.

The majority of researchers and practitioners prefer rather low rate of
mutations.  This is because the very frequent mutations turn the
genetic algorithm into generic Monte Carlo procedure, which is to be
avoided, since we hope that some kind of intelligence will lead us {\em
much\/} faster to the desired solution than completely blind trials.

\section{The fate of mutated genes}
Suppose, for simplicity, that the smallest part of the chromosome,
called here gene, consists of only a single bit.  Let this bit has
initial value of $0$.  The mutation process is then nothing else than
negating (flipping) this bit.  The question we want to examine now is:
\begin{quote}
	\sl What will be the state of this bit after $k$ generations?
\end{quote}
Let the probability of bit flip per generation is equal to $p$.  We can
write
\begin{equation}\label{master_eq}
	\begin{array}{rclcl}
	p_{k+1}(1) &=& p_{k}(0)\cdot p &+& p_{k}(1)\cdot (1-p)\\
	p_{k+1}(0) &=& p_{k}(0)\cdot (1-p) &+& p_{k}(1)\cdot p
	\end{array}
\end{equation}
where $k$ numerates consecutive generations and the number in
parenthesis means the state of the bit in question.  Taking into
account that for any $k$
\begin{equation}
	p_{k}(0) + p_{k}(1) = 1
\end{equation}
we can rewrite the first row of relation (\ref{master_eq}) as
\begin{eqnarray}
	p_{k+1}(1) &=& \left[1-p_{k}(1)\right] p + p_{k}(1)\cdot (1-p)\\
	&=& p_{k}(1) \left(1 - 2 p\right) + p
\end{eqnarray}
or, in shorter form
\begin{equation}\label{recur}
	p_{k+1} = (1-2p)p_{k}+p
\end{equation}
where $p_{j}$ is the probability that the examined bit is in the state
$1$ after $j$ generations.

The recurrence formula (\ref{recur}) seems very similar to the
construction of a simple random number generator, so one might expect
that it produces chaotic sequences of numbers.  Surprisingly (?) this
is not the case, as we shall see.

Consider the behavior of $p_{k}$ when $k\rightarrow\infty$.  It is easy
to see, that for $p=0$ (no mutations) the sequence $p_{k}$ is constant
(and thus convergent); it remains at whatever initial state -- in our
case $0$.  The other limiting case, $p=1$, is also easy: now the
recurrence relation (\ref{recur}) simplifies to
\begin{equation}
	p_{k+1} = 1 - p_{k}
\end{equation}
and our bit permanently oscillates between two possible states
$0\leftrightarrow 1$ --- there is no convergence.  For all other cases,
i.e. $0<p<1$, the sequence (\ref{recur}) is always convergent and we
have
\begin{equation}
	\lim_{k \rightarrow\infty} p_{k} = \frac{1}{2}
\end{equation}
We skip the easy proof.  It is interesting, however,  that the
convergence has oscillatory character when $p>\frac{1}{2}$, while for
$p<\frac{1}{2}$ the sequence $p_{k}$ grows monotonically.  There is no
chaotic behavior. The case $p=\frac{1}{2}$ is special again: it
produces constant sequence $p_{k}=\frac{1}{2}$.

\section{Aging of the population}

It is widely believed, that the mutations are essentially wrong thing
for any individual.  They usually damage the genetic material,
sometimes to the extent which makes further existence of the individual
impossible.  It only rarely happens that the mutated individual is
better fitted to its environment than the average one in a given
population.  On the other hand not all mutations are lethal.  In the
well known Penna \cite{Penna1}, \cite{Penna2} model of biological
evolution, for example, the subsequent mutations are accumulated in
every chromosome.  The older the chromosome, the more mutations it
carries and finally dies either after single lethal mutation or due to
the accumulation of many less damaging defects (we do not discuss here
the so called Verhulst factor describing the decrease of the population
size due to overcrowding and thus the limited accesss to necessary
resources by any single individual).  If no replication occurs, then
the entire population is eventually extinct.  By contrast, the
``small'' changes in the genotypes are silently transmitted to the
offspring chromosomes.

In the Nature, however, an error correcting capability is at work.  The
double helix structure of the DNA strand limits the proliferation of
defective genes. The two parts of DNA (and RNA as well) are not
independent, but complementary. After unzipping, just before the
replication, only one strand of DNA contains the distorted genetic
information, while the other sequence is correct.  Therefore only
$50$\% of offspring DNA helices will be damaged.

\section{Implementation in genetic algorithms}

The idea is to use in genetic algorithms the chromosomes, which are
``doubled'', i.e. which consist of regular, well known part
(``visible'') and the other (``invisible'') parallel structure, of
identical length, which is initially the exact negation of the first
part, as below:
\begin{equation}
\begin{tabular}{|c|c|c|c|c|c|c|c|c|c|c|c|c|c|c|c|c|l}
\cline{1-17}
	0&0&1&0&0&1&1&0& \ldots &1&1&1&0&1&0&0&1 & --- visible part\\
\cline{1-17}
	1&1&0&1&1&0&0&1& \ldots &0&0&0&1&0&1&1&0 & --- invisible part\\
\cline{1-17}
\end{tabular}
\end{equation}
We do not mark any gene boundaries here.  A single gene may consist of
one or more bits; the lengths of consecutive genes need not to be equal
to each other.  The chromosome becomes mutated when at least one of its
bits is flipped.

The genetic material, representing trial solution of the original
problem, is placed in the visible part.  Every mutation is reflected in
the state of this very part, the invisible part being completely
insensitive to mutations.  Quite contrary, the crossover operation is
performed on both parts, mixing independently two visible and two
invisible parts of the involved chromosomes, at the same crossover
point(s).  It is easy to see, that the crossover operation alone, no
matter one- or multipoint, will never destroy the complementarity of
the visible and invisible part of any chromosome.  Of course, the
fitness factor will always be computed for the visible part only. Now
the process of aging of the population, generation after generation,
may be tracked by looking at the invisible part of every chromosome. 
We can easily count all the bits in the entire population, which were
changed by mutations --- it is enough to compare the visible and
invisible parts of all chromosomes and count the bits, which are
identical in both parts.  Of course, the even number of mutations
applied to the same bit will go unnoticed.  This is sometimes called
{\em backward evolution\/}.  From the earlier considerations, and
assuming that the mutation probability $p$ per bit and per generation
is small (more precisely $p<\frac{1}{2}$, as is always the case), we
can conclude that the fraction of (effectively) mutated bits in the
entire population will be an increasing function of time.  The
stochastic limit for this fraction is equal to $\frac{1}{2}$.  The
value of $p=\frac{1}{2}$ uniquely separates the genetic algorithms from
Monte Carlo type of optimization.

\section{Discussion}
For $p \ll \frac{1}{2}$ the fraction of mutated bits initially
increases linearly with the number of generations.  This observation
does not take into account directly neither the structure of the
individual chromosome (number of genes or bits it consists of) nor the
number of chromosomes in the population.  Such information is hidden in
the value of parameter $p$ --- probability of a particular bit being
flipped during a~single cycle of simulation (single epoch). Thus the
the number of generations to reach $50$\% of flipped bits may be
roughly estimated as $N_{gen} = \frac{1}{2p}$.  In practice, as the
performed simulations show, the fraction of effectively mutated bits
reaches the value of $\frac{1}{2}$ --- for the first time --- after
roughly $5$ to $6$ times greater number of epochs.  The exact
analytical formula for $p_{k}$ as a function of $k$ is neither easy to
obtain nor very informative.  For any $k$, $p_{k}$ is a polynomial of
order $k$ in variable $p$.  The graph of $p_{k}$ vs. $k$, however, shows
striking similarity to the graph of the expression:
\begin{equation}
	\frac{1}{2}\left[ 1 -\exp \left(-2kp\right) \right]
\end{equation}
This is only an observation based on several evolutionary processes,
guessed rather than formally derived.  Anyway, utilizing this
approximation, we can say that after $\frac{1}{2p}$ epochs the
population is mature and gradually looses its innovative forces, and
after
\begin{equation}\label{result}
	N_{gen} = \frac{3}{p}
\end{equation}
epochs it becomes old and practically useless.  It is time then to
switch to other locally searching routine to improve further the
optimal set of unknown parameters, if appropriate.  The formula
(\ref{result}) is the main result of this paper.  It sets the {\em
upper\/} limit for the number of necessary generations.

In conclusion, during calculations we should monitor the behavior of
the fraction of mutated bits.  {\bf When this variable reaches the
value of} $\frac{1}{2}$, {\bf then we may say that every second bit in
the population was flipped at least once.  This is the other way of
saying that the search space has been explored quite thoroughly and no
significant improvement of the fitness should be expected}.

\medskip
{\sl Why do we make such a claim?}

\medskip
If the fitness of all chromosomes was the same, i.e. there was no
evolutionary pressure of any kind, then after $\frac{3}{p}$ generations
the search space would be covered quite uniformly, although
irregularly, with trial points.  Further sampling would only make finer
the already existing irregular grid of trial points.  When there {\em
is\/} a preference for better fitted chromosomes, then after
$\frac{1}{2p}$ generations all interesting regions, or at least their
neighborhoods, should have already been found.  Observing the values of
the fitness function alone, either as an average for the population or
for the best individual only, may be {\em very\/} misleading.

\medskip
Considering the efficiency of the algorithm, understood as the number
of generations necessary to find the optimal fitness, we can see that
it is inversely proportional to $p$.  For every practical purpose the
condition
\begin{equation}\label{min-mut}
	pN_{bits}=1
\end{equation}
(average number of bits flipped in the whole population during one
epoch is equal to one) sets the lowest sensible limit for $p$.
$N_{bits}$ is the total number of bits in the population (in its
visible part).  Of course, higher values of $p$ will speed up the
evolutionary search.

\medskip
{\sl Is the prescription}~ ~\fbox{$N_{gen} \approx\frac{1}{2p}$}~ ~{\sl
perfect?}

\medskip
Certainly not.  It gives the reasonable estimate of the number of
generations needed in fairly regular cases.  The genotype space with
only one well fitted chromosome and all other equally bad is an evident
exception.  What can be done in such cases is to increase the number of
bits assigned to every continuous unknown (kind of {\em
oversampling\/}), for the price of increased computational effort, of
course, since this approach is equivalent to the use of finer grid of
points in the search space. To purely combinatorial problems (only
integer and/or logical unknowns) our estimate should be of similar
value.

\medskip
As an example let us take the data from \cite{III_KK}.  Here the
population consisted of 30 chromosomes, $20$ bits long each.  and $p$
was set equal to $0.03$.  According to (\ref{min-mut}), $p$ should be
set as at least $1/(30\times 20) \approx 0.0017$ --- while the quoted
mutation rate was $18$ times higher than that number.  The population
should become mature after some $1/(2\times 0.03)\approx 17$
generations and the evolutionary search should be terminated after
$100$ generations at the latest.  In~fact, the author of \cite{III_KK}
reports that satisfactory results were achieved after $10$---$19$
generations (every run was limited to $20$ generations).  The search
space had only $2^{20}=1~048~576$ points, so this example may be
considered small and not representative.  Nevertheless, no more than
only $600$ evaluations of the fitness factor were enough to reach
valuable conclusions.

\section{Why is genetic algorithm so effective anyway?}
We must be aware, that the search space, no matter how large, is always
{\em discrete\/} and finite for this class of optimization algorithms. 
It may be considered as a random graph, in which every chromosome is a
vertex (we are not interested in the edges).  This graph is by no means
random, but clearly exhibits the {\em small world property\/}, i.e. the
average (Hamming) distance between its vertices scales as the logarithm
of their number.  Indeed, $N$-bit chromosomes are points in $2^{N}$
element universe.  The maximum distance between any two chromosomes is
equal to $N$, so the average distance must never exceed this number. 
Incidentally $\log_{2}2^{N}=N$.  This fact alone explains why genetic
algorithms are fairly insensitive with respect to the number of
unknowns.  On the other hand, speaking of the neighboring points in the
search space makes sense, especially when we think of {\em nearest
neighbors\/}.  It is therefore intuitively appealing that we should be
able to find a (hopefully small) subset of points in the search space
with the property that {\em any\/} point is distant from this set no
more than $1$ unit -- something similar to the backbone or spanning
tree known from the graph theory.  In case $N=3$ it is easy to check
that exactly two points, namely $\left(000\right)$ and
$\left(111\right)$, are enough to form such a subset with requested
property.  Evaluating fitness for each member of this subset should be
nearly equivalent to the exhaustive (``brute force'') search, since
then, for arbitrarily chosen point in the search space either this
point itself or one of its nearest neighbors was visited and evaluated
during evolutionary process.  Unfortunately, today we don't know how to
find such a set in general case; we don't even know what is its
cardinality -- hopefully much lower than that of original search space.
It is certain, however, that the solution of this problem need not to
be unique.\footnote{Consider the simple case of
$N=2$ -bit chromosomes.  The universe consists of just $4$ elements:
$\left(00\right)$, $\left(01\right)$, $\left(10\right)$ and
$\left(11\right)$.  There are two subsets with desired property:
$\left\{\left(01\right), \left(10\right)\right\}$ and
$\left\{\left(00\right), \left(11\right)\right\}$.}
Perhaps the genetic algorithm is the best available tool for
approximate construction of such subsets?

\section{Comments on performance}
Using ``doubled'' chromosomes we successfully mimic the double helix
structure of DNA.  The cost is moderate: we need to double the storage
for the population.  Counting the flipped bits is performed only once
per generation, so this cost should be negligible in comparison with
evaluations of fitness function.  Instead of the true implementation of
the ``double helix'', one can use the simplified formula (\ref{result})
as a stopping criterion.  Direct observation of the fraction of
effectively mutated bits will signal the end of calculations usually
much earlier.

\section{Acknowledgment}
This work was done as part of author's statutory activity in Institute
of Physics, Polish Academy of Sciences.

%%%%% bibliography
%%%%% %%%%%%%%%%%%%%%%%%%%%%%%%%%%%%%%%%%%%%%%%%%%%%%%%%%%

\end{document}